\documentclass[a4paper]{spie}  %>>> use this instead for A4 paper

\usepackage{amsmath,amsfonts,amssymb}
\usepackage{graphicx}
\usepackage[colorlinks=true, allcolors=blue]{hyperref}
\usepackage{siunitx}

\usepackage{threeparttable}

\title{Response monitoring of breast cancer on DCE-MRI using convolutional neural network-generated seed points and constrained volume growing}

\author[1]{Bas H.M. van der Velden}
\author[1]{Bob D. de Vos}
\author[2]{Claudette E. Loo}
\author[1]{Hugo J. Kuijf}
\author[1]{Ivana I\v{s}gum}
\author[1]{Kenneth G.A. Gilhuijs}
\affil[1]{Image Sciences Institute, University Medical Center Utrecht, Utrecht, The Netherlands}
\affil[2]{Department of Radiology, Antoni van Leeuwenhoek Hospital - The Netherlands Cancer Institute, Amsterdam, The Netherlands}

\begin{document}
\maketitle

\begin{abstract}

Response of breast cancer to neoadjuvant chemotherapy (NAC) can be monitored using the change in visible tumor on magnetic resonance imaging (MRI). In our current workflow, seed points are manually placed in areas of enhancement likely to contain cancer. A constrained volume growing method uses these manually placed seed points as input and generates a tumor segmentation. This method is rigorously validated using complete pathological embedding. In this study, we propose to exploit deep learning for fast and automatic seed point detection, replacing manual seed point placement in our existing and well-validated workflow. The seed point generator was developed in early breast cancer patients with pathology-proven segmentations (N=100), operated shortly after MRI. It consisted of an ensemble of three independently trained fully convolutional dilated neural networks that classified breast voxels as tumor or non-tumor. Subsequently, local maxima were used as seed points for volume growing in patients receiving NAC (N=10). The percentage of tumor volume change was evaluated against semi-automatic segmentations. The primary cancer was localized in 95\% of the tumors at the cost of 0.9 false positive per patient. False positives included focally enhancing regions of unknown origin and parts of the intramammary blood vessels. Volume growing from the seed points showed a median tumor volume decrease of 70\% \mbox{(interquartile range: 50\%--77\%)}, comparable to the semi-automatic segmentations \mbox{(median: 70\%, interquartile range 23\%--76\%)}. To conclude, a fast and automatic seed point generator was developed, fully automating a well-validated semi-automatic workflow for response monitoring of breast cancer to neoadjuvant chemotherapy.
\end{abstract}


\section{INTRODUCTION}
Treatment of breast cancer consists of surgery, often followed by radiotherapy and systemic therapy. Systemic therapy --- and particularly chemotherapy --- can be administered prior to surgery. This so-called neoadjuvant chemotherapy (NAC) allows shrinking of the tumor before surgery, reducing surgical extent, complications, and improving cosmetic outcome \cite{Fisher1997,Mauri2005,Mieog2007}. \let\thefootnote\relax\footnote{This work has been accepted for SPIE Medical Imaging 2019, Computer-Aided Diagnosis conference, Paper 10950-12.}

The response of the tumor to NAC can be monitored by imaging the tumor \cite{Steenbruggen2017}. The change in visible tumor volume on dynamic contrast-enhanced magnetic resonance imaging (DCE-MRI) can be used as a measure of response \cite{Hylton2012}. Consequently, the NAC treatment plan can be adjusted when tumor response is deemed insufficient. 

In our current response monitoring workflow, seed points are manually placed in areas of enhancement that are likely to contain tumor. A constrained volume growing method uses these manually placed seed points as input and generates a tumor segmentation \cite{Gilhuijs2002}. This method was rigorously validated using complete pathological embedding \cite{Alderliesten2007}. Even though the method is very accurate \cite{Alderliesten2007}, manual placement of the required seed points is a time consuming step, prone to intraobserver and interobserver variability. Therefore, automating placement of these seed points is of interest. 

Deep learning algorithms have shown high robustness in a variety of medical image analysis tasks \cite{Litjens2017}. In breast MRI, these methods have mainly focused on diagnosis or on segmentation of healthy tissues \cite{Dalms2017,Moeskops2016,Amit2017}. In this study, we propose to exploit deep learning for fast and automatic seed point detection, replacing manual seed point placement in our existing and well-validated workflow \cite{Gilhuijs2002,Alderliesten2007}. 

\section{MATERIAL AND METHODS}

We developed the automatic seed point generator using an ensemble of three independently trained fully convolutional dilated neural networks. Two patient cohorts were used in the study: Cohort 1 consisted of early breast cancer patients in which the seed point detector was developed and evaluated, Cohort 2 consisted of patients receiving NAC in which seed points were automatically placed for constrained volume growing. Tumor response to NAC was evaluated by comparing the change in tumor volume to our reference standard at the MRI before NAC and the MRI during NAC. This reference standard consisted of semi-automatically obtained tumor segmentations. The study design is illustrated in Figure 1, the following paragraphs describe these steps in detail.

\begin{figure}[h]
\centering
\includegraphics[width=1.0\textwidth]{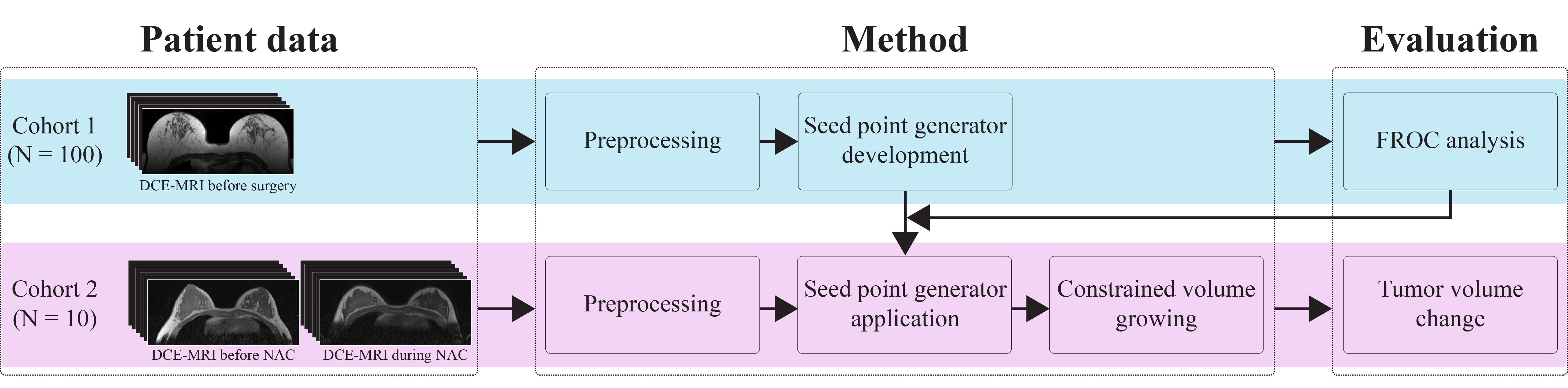}
\caption{Overview of the study. FROC = free-response receiver operating characteristics.}
\end{figure}

\subsection{Patients}

\textbf{Cohort 1:} The seed point generator was developed in a subset of 100 randomly selected patients with invasive ductal carcinoma from the prospective Multimodality Analysis and Radiological Guidance IN breast conServing therapy study (MARGINS, 2000-2008) \cite{Elshof2010}. Patients who were eligible for breast conserving therapy on the basis of conventional imaging and clinical assessment were consecutively recruited for an additional preoperative breast MRI. 

\textbf{Cohort 2:} The constrained volume growing was applied to generated seed points in 10 patients with invasive breast cancer who received NAC who particitpated in a prospective studie (2008-2013) \cite{Loo2011}. Eligibility criteria included primary invasive breast cancer of \mbox{3 cm} or larger and/or one or more tumor-positive axillary lymph node. 

The baseline characteristics of the patient cohorts are displayed in Table 1.

\begin{table*}[t]
\caption{Baseline characteristics of the patient cohorts}
\begin{center}
\renewcommand{\arraystretch}{1.0}
\begin{tabular}{lccr}

\hline
\textbf{Characteristics}  &  \textbf{Cohort 1 (N = 100)}& \textbf{Cohort 2 (N = 10)} & \textbf{P-value$^\ast$}    \\
\hline

Age (years)$^\dagger$	 & 57 (32 -- 84) & 42 (29 -- 64) & 0.009  \\
Largest tumor diameter on MRI (mm)$^\dagger$ & 21 (9 -- 38) & 37 (25 -- 59) & $<$ 0.001     \\
Immunohistochemical subtype		& 		 &		 & 0.002\\
\hspace*{0.25cm}ER-positive/HER2-negative 	& 73 (73)& 4 (40)&\\
\hspace*{0.25cm}HER2-positive			& 15 (15)& 0 (0)&\\
\hspace*{0.25cm}Triple negative & 12 (12)& 6 (60)&\\
Histologic grade 				&		 &		 & 0.001 \\
\hspace*{0.25cm}Grade I 		& 30 (30)& 0 (0) &\\
\hspace*{0.25cm}Grade II 		& 39 (39)& 1 (10)&\\
\hspace*{0.25cm}Grade III 		& 31 (31)& 9 (90)&\\
Tumor-positive axillary lymph nodes & 		 &		 & 0.116 \\
\hspace*{0.25cm}0  				& 61 (61)& 3 (30)&\\
\hspace*{0.25cm}1--3 			& 31 (31)& 6 (60)&\\
\hspace*{0.25cm}4 or more    	& 8 (8)  & 1 (10)&\\
\hline

\end{tabular}

\begin{tablenotes}
\begin{small}
Values are number of patients with percentages in parentheses, unless specified otherwise.\\
$^\dagger$ Values are medians, with ranges in parentheses.\\
$^\ast$ Significance levels for differences between patient cohorts are calculated using the Kruskal-Wallis test for \\continuous variables and using Fisher's exact test for categorical variables.\\
ER = estrogen receptor, HER2 = human epidermal growth factor receptor 2.
\end{small}
\end{tablenotes}

\end{center}
\end{table*}

\subsection{Magnetic resonance imaging}
\textbf{Cohort 1:} The patients received one MRI --- before surgery. MR images were acquired using a 1.5 T imaging unit (Magnetom; Siemens, Erlangen, Germany) with a dedicated double breast array coil. Five T\textsubscript{1}-weighted images were acquired: one precontrast image and four postcontrast images after injection of gadolinium-based contrast material (Prohance; Bracco-Byk Gulden, Konstanz, Germany). The imaging parameters were: repetition time 8.1 ms, echo time 4.0 ms, flip angle \ang{20}, isotropic voxel size 1.35 $\times$ 1.35 $\times$ 1.35 mm\textsuperscript{3}, and field of view 310 mm.

\textbf{Cohort 2:} The patients received an MRI before NAC and one after six weeks of NAC. MR images were acquired using a 3 T imaging unit (Achieva; Philips Medical Systems, Best, the Netherlands) with a 7-elements sense breast coil. Six T\textsubscript{1}-weighted images were acquired: one precontrast image and five postcontrast images after injection of gadolinium-based contrast material (Prohance; Bracco-Byk Gulden, Konstanz, Germany). The imaging parameters were: repetition time 4.4 ms, echo time 2.3 ms, flip angle \ang{10}, isotropic voxel size \mbox{1.1 $\times$ 1.1 $\times$ 1.1 mm\textsuperscript{3}}, and field of view 360 mm.

\subsection{Method}
Preprocessing consisted of deformable registration between the precontrast MRI and the postcontrast images \cite{Dmitriev2013}, and automatic segmentation of the breast area in three dimensions \cite{VanderVelden2015}.

\begin{figure}[b]
\centering
\includegraphics[width=1.0\textwidth]{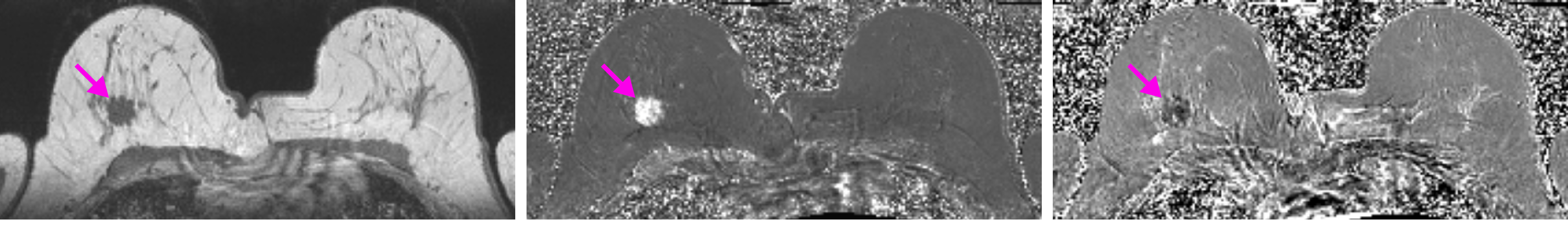}
\caption{Example of the input images. Left: Transversal precontrast T\textsubscript{1}-weighted image of a 58 year old patient with a 2.1 cm \mbox{grade 2} invasive ductal carcinoma, the tumor is indicated by the pink arrow. Middle: The corresponding washin image, clearly visualizing the tumor. Right: The corresponding washout image.}
\end{figure}

Seed point detection was achieved using a dilated convolutional neural network. The convolutional neural network analyzed two-dimensional images with three channels (Figure 2). The first channel consisted of the precontrast image to feed anatomical knowledge to the system. The precontrast images were corrected for field inhomogeneities\cite{Tustison2010}, and normalized between zero and one on the 5\textsuperscript{th} and 95\textsuperscript{th} intensity percentiles under the breast mask. The second channel consisted of the washin: the relative change in image intensities between the precontrast and the first postcontrast scan \cite{Gilhuijs1998}. The third channel consisted of the washout: the relative change in image intensities between the first postcontrast and the last postcontrast scan \cite{Gilhuijs1998}. The washin and washout are by definition normalized measures \cite{Gilhuijs1998}. Tumors tend to show high washin and high negative washout \cite{Morris2013}.

The architecture of the network is shown in Figure 3. The network had a receptive field of \mbox{97 $\times$ 97} voxels \cite{Yu2016}. Due to increasing dilation factors in subsequent convolutional layers, our network with nine two-dimensional convolutional layers resulted in only $57\,506$ trainable weights. Rectified linear units were used for activation, except for the final layer, which had a binary softmax output. Because the network was fully convolutional, it could be applied to images of arbitrary size \cite{Wolterink2017}. Of the 100 patients, 60 were used for training of the network, 20 for validation, and 20 for independent testing.

\begin{figure}[t]
\centering
\includegraphics[width=1.0\textwidth]{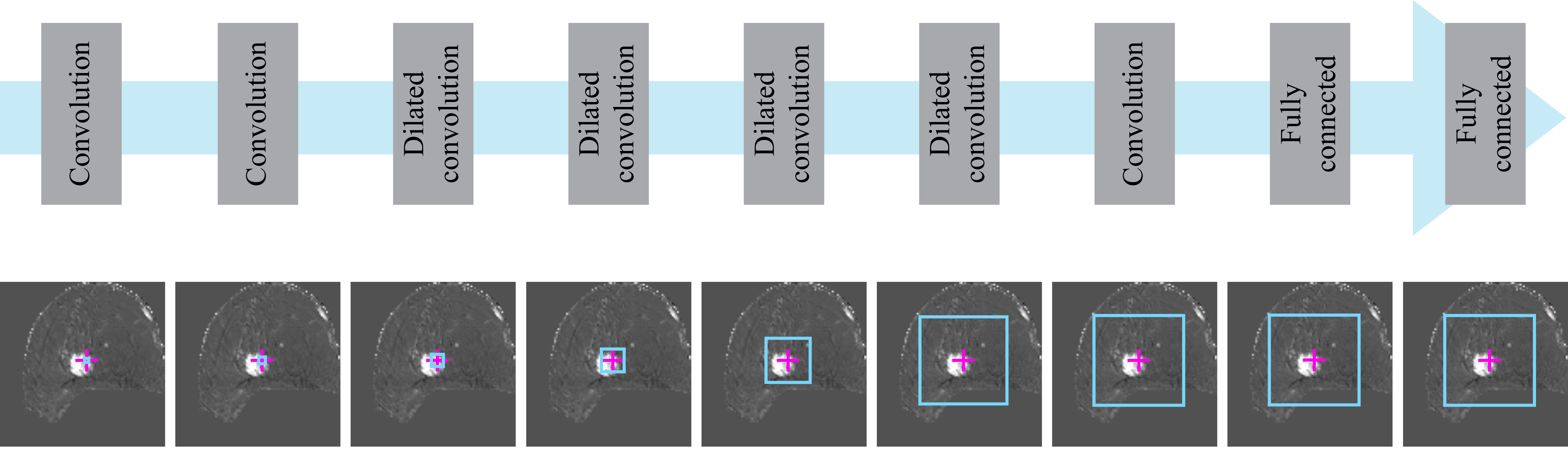}
\caption{Top: Architecture of the dilated convolutional neural network \cite{Yu2016}. The convolution layers have 32 3 $\times$ 3 filters followed by a rectified linear unit activation function. The last two layers consist of fully connected layers (implemented as 1 $\times$ 1 convolutions), of which the final layer is followed by a softmax function. Bottom: Visualization of the receptive field at each layer \cite{Wolterink2017}. A part of the transversal washin image masked by the breast mask is shown for each convolution layer. The blue line illustrates the receptive field influencing the classification of the voxel indicated by the pink cross hair.}
\end{figure}

An ensemble of three different random initializations of the above-described network was trained in $100\,000$ minibatches. Minibatches consisted of 25 patches of 99 $\times$ 99, equally balanced to contain tumor or not. The best model was chosen based on the validation loss between the $50\,000$\textsuperscript{th} and $100\,000$\textsuperscript{th} iteration. Each network was trained with a different order of the training images. During application of the trained networks, posterior probabilities were thresholded and consensus voting was used on the resulting segmentation maps. Consensus results were used to generate seed points by applying an inward Euclidean distance transform and subsequent local maxima extraction.

Finally, tumor segmentation was performed using our constrained volume growing method that has been validated using complete pathological embedding \cite{Alderliesten2007}. In short, starting from a seed point in or near the tumor, the method enhanced the tumor, constructed an ellipsoidal volume of interest around the tumor, and applied a voxel-value based stopping criterion automatically derived from the enhancing tumor \cite{Alderliesten2007}.

\subsection{Evaluation}
The detection performance of the automatic seed point generator was evaluated in Cohort 1 using free-response receiver operating characteristics. The seed point generator was subsequently applied to Cohort 2 to assess its application in a response monitoring workflow. The change in tumor volume between the MRI before NAC and the MRI during NAC was calculated, and compared with our reference standard that consisted of segmentations acquired by constrained volume growing from manually placed seed points.

\section{RESULTS}

\subsection{Cohort 1: Evaluation of seed point generator}
At a threshold on the posterior probability of 0.9, the model with the consensus voting localized 95\% of the lesions for the independent test set at a cost of 0.9 false positives per patient. Seed point generation took approximately \mbox{0.7 second} per patient.

\begin{figure}[t]
\centering
\includegraphics[width=1.0\textwidth]{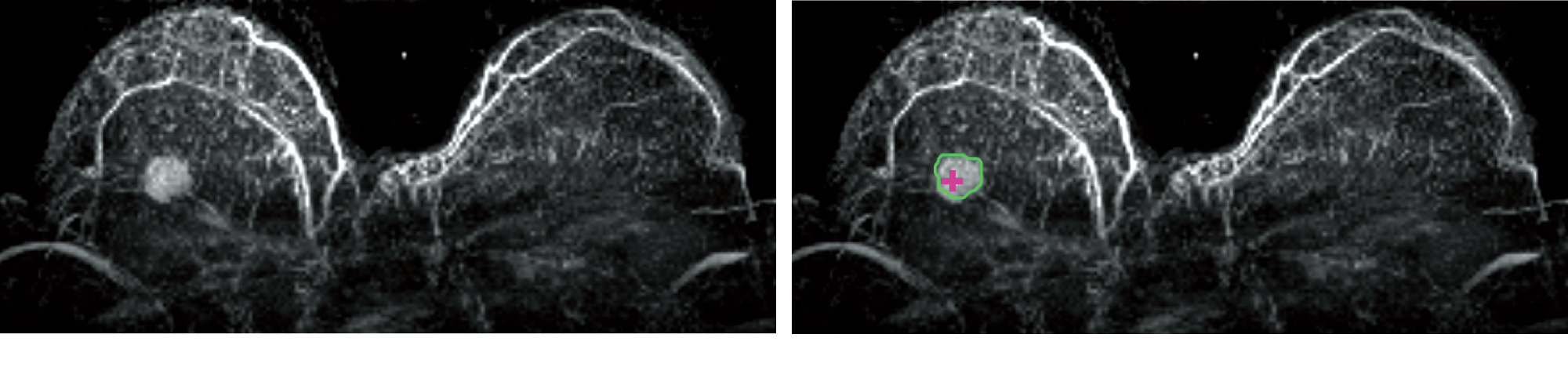}
\caption{Example of automatic seed point generation in a patient from the independent test set. Transversal maximum intensity projection of a subtraction image showing a tumor in the right breast (left for viewer). The image on the right shows the ground truth with the green contour, and the seed point after consensus voting of the model with the pink crosshair.}
\end{figure}

\subsection{Cohort 1: Inspection of false negatives and false positives}
We inspected the false negatives and false positives of the model after the consensus voting. The primary tumor was missed in one of the patients from the independent test set, this tumor was found by just one of the three network initializations. The false positives were focally enhancing regions --- potentially indicative of benign disease (N = 4 patients), part of an intramammary blood vessel (N = 1 patient), and seemingly healthy enhancing parenchyma (N = 2 patients).

\subsection{Cohort 2: Response to neoadjuvant chemotherapy}
The constrained volume growing after consensus voting showed a median decrease in tumor volume of 70\% (interquartile range: 50\% -- 77\%), comparable to the reference standard \mbox{(70\%, interquartile range: 23\% -- 76\%)} (Figure 3). Inspection of the segmentations suggests that the method segments intramammary blood vessels more often in the patients receiving NAC (3/10, 30\%) than in the patients with early breast cancer (1/20, 5\%). Seed point generation took approximately \mbox{1.4 second} per patient.

\begin{figure}[h]
\centering
\includegraphics[width=1.0\textwidth]{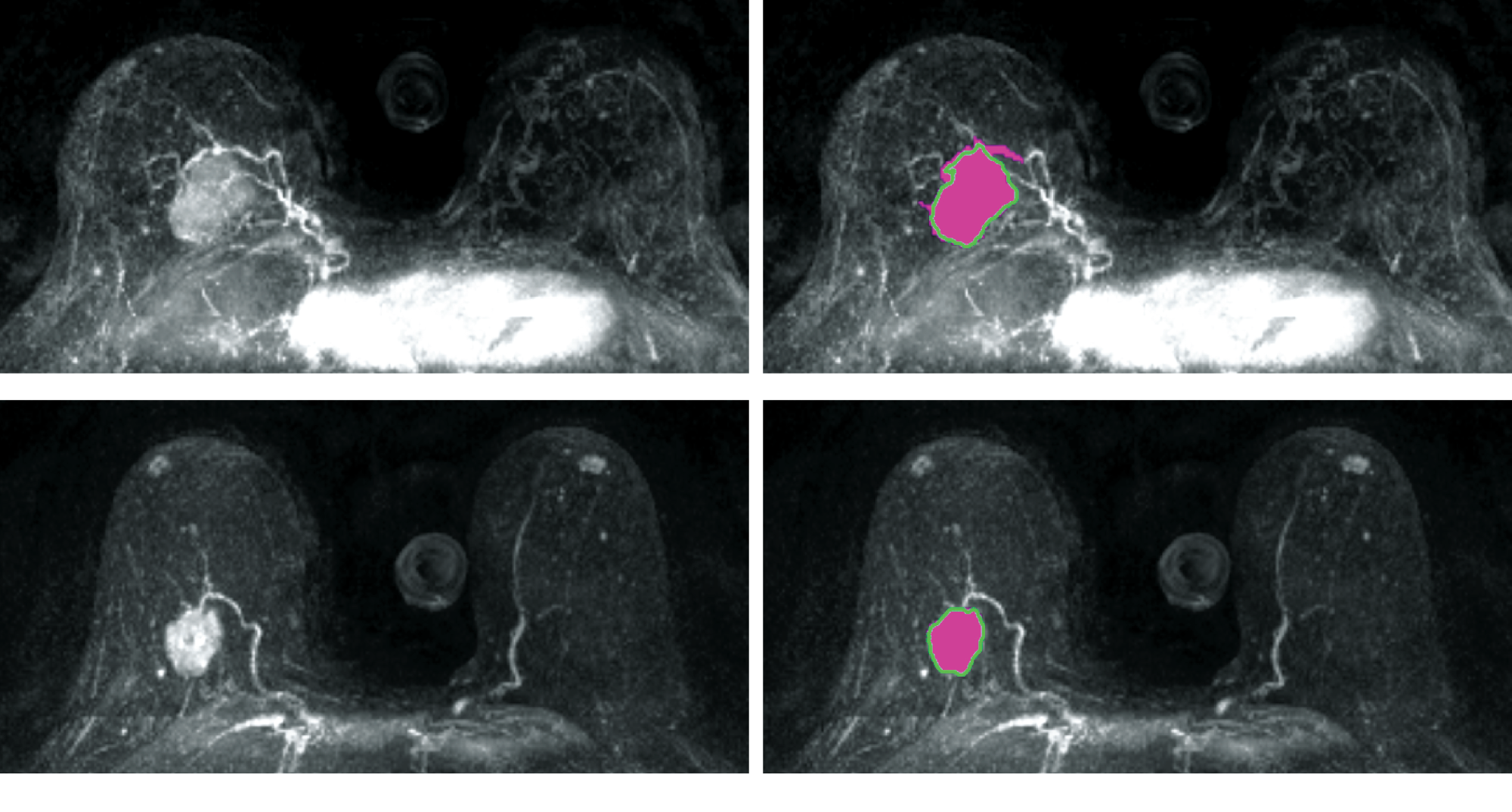}
\caption{Example of tumor shrinkage by neoadjuvant chemotherapy. Top row: Transversal maximum intensity projection of a subtraction image showing a tumor in the right breast. The image on the right shows the reference standard with the green contour and the output of the method with the pink overlay. Bottom row: Scan after six weeks of chemotherapy. Notice that the tumor has shrunk, represented by both the reference standard (volume decrease of 69\%) and the proposed method (volume decrease of 70\%).}
\end{figure}

\section{DISCUSSION}
We presented a fully automated method for response monitoring to neoadjuvant chemotherapy (NAC). The method showed tumor shrinkage comparable to our reference standard. Hence, these preliminary results suggests that this method is feasible in a response monitoring workflow.

The method replaced a manual step in our workflow for response monitoring to NAC. Thus far, seed points were manually placed, after which a constrained volume growing algorithm segmented the tumor. In this study, we automatically calculated these seed points in roughly a second per patient. The primary tumor was located in 95\% of the patients at the cost of 0.9 false positives per patient --- comparable to literature \cite{Gubern-Merida2015}. The seed point generator was applied as input for the constrained volume growing in patients receiving NAC. The change in tumor volume between the MRI before NAC and the MRI during NAC (median decrease 70\%, \mbox{interquartile range: 50\% -- 77\%)} was comparable to our reference standard (median decrease 70\%, \mbox{interquartile range: 23\% -- 76\%)}.

False positives mainly occur in the intramammary blood vessels. Therefore, future work could investigate suppressing the blood vessels, either by preprocessing\cite{Vignati2014,VanDerVelden2016}, or by segmenting blood vessels as a separate class. That way, the network can potentially better distinguish them.

Regardless of these shortcomings, the method does perform comparable to our reference standard. Note that the method is trained on a different patient population, with MR images from a different vendor (Philips instead of Siemens), and differences in MR acquisition (3T vs 1.5T, slightly different voxel size, different repetition time, echo time, and flip angle). Hence, the method appears to be relatively agnostic to these patient and imaging differences.

In conclusion, response monitoring of breast cancer to neoadjuvant chemotherapy on dynamic contrast-enhanced MRI using convolutional neural network-generated seed points and constrained volume growing is feasible. 

\section{NEW OR BREAKTHROUGH WORK}
A fast and automatic seed point generator was developed, fully automating a well-validated semi-automatic workflow for response monitoring of breast cancer to neoadjuvant chemotherapy.

\section{ACKNOWLEDGMENTS}
This work is partially funded by the Dutch Cancer Society (KWF), grant number 10755, and by the Center for Translational Molecular Medicine (CTMM) grants Breast CaRe and CHOICE. We thank NVIDIA for the generous donation of a GPU.

\bibliography{mendeley2}

\begin{thebibliography}{10}

\bibitem{Fisher1997}
Fisher, B., Brown, A., Mamounas, E., Wieand, S., Robidoux, A., Margolese,
  R.~G., Cruz, A.~B., Fisher, E.~R., Wickerham, D.~L., Wolmark, N., DeCillis,
  A., Hoehn, J.~L., Lees, A.~W., and Dimitrov, N.~V., ``{Effect of preoperative
  chemotherapy on local-regional disease in women with operable breast cancer:
  findings from National Surgical Adjuvant Breast and Bowel Project B-18},''
  {\em J. Clin. Oncol.}~{\bf 15}(7),  2483--2493 (1997).

\bibitem{Mauri2005}
Mauri, D., Pavlidis, N., and Ioannidis, J. P.~A., ``{Neoadjuvant Versus
  Adjuvant Systemic Treatment in Breast Cancer: A Meta-Analysis},'' {\em JNCI
  J. Natl. Cancer Inst.}~{\bf 97}(3),  188--194 (2005).

\bibitem{Mieog2007}
Mieog, J. S.~D., Van~der Hage, J.~A., and Van~de Velde, C. J.~H.,
  ``{Neoadjuvant chemotherapy for operable breast cancer.},'' {\em Br. J.
  Surg.}~{\bf 94}(10),  1189--200 (2007).

\bibitem{Steenbruggen2017}
Steenbruggen, T.~G., Van~Ramshorst, M.~S., Kok, M., Linn, S.~C., Smorenburg,
  C.~H., and Sonke, G.~S., ``{Neoadjuvant Therapy for Breast Cancer:
  Established Concepts and Emerging Strategies.},'' {\em Drugs}~{\bf 77}(12),
  1313--1336 (2017).

\bibitem{Hylton2012}
Hylton, N.~M., Blume, J.~D., Bernreuter, W.~K., Pisano, E.~D., Rosen, M.~A.,
  Morris, E.~A., Weatherall, P.~T., Lehman, C.~D., Newstead, G.~M., Polin, S.,
  Marques, H.~S., Esserman, L.~J., and Schnall, M.~D., ``{Locally Advanced
  Breast Cancer: MR Imaging for Prediction of Response to Neoadjuvant
  Chemotherapy—Results from ACRIN 6657/I-SPY TRIAL},'' {\em Radiology}~{\bf
  263}(3),  663--672 (2012).

\bibitem{Gilhuijs2002}
Gilhuijs, K. G.~A., Deurloo, E.~E., Muller, S.~H., Peterse, J.~L., and
  {Schultze Kool}, L.~J., ``{Breast MR Imaging in Women at Increased Lifetime
  Risk of Breast Cancer: Clinical System for Computerized Assessment of Breast
  Lesions—Initial Results},'' {\em Radiology}~{\bf 225}(3),  907--916 (2002).

\bibitem{Alderliesten2007}
Alderliesten, T., Schlief, A., Peterse, J., Loo, C., Teertstra, H., Muller, S.,
  and Gilhuijs, K., ``{Validation of semiautomatic measurement of the extent of
  breast tumors using contrast-enhanced magnetic resonance imaging.},'' {\em
  Invest. Radiol.}~{\bf 42}(1),  42--9 (2007).

\bibitem{Litjens2017}
Litjens, G., Kooi, T., Bejnordi, B.~E., Setio, A. A.~A., Ciompi, F.,
  Ghafoorian, M., Van~der Laak, J.~A., Van~Ginneken, B., and S{\'{a}}nchez,
  C.~I., ``{A survey on deep learning in medical image analysis},'' {\em Med.
  Image Anal.}~{\bf 42},  60--88 (2017).

\bibitem{Dalms2017}
Dalmış, M.~U., Litjens, G., Holland, K., Setio, A., Mann, R., Karssemeijer,
  N., and Gubern-M{\'{e}}rida, A., ``{Using deep learning to segment breast and
  fibroglandular tissue in MRI volumes},'' {\em Med. Phys.}~{\bf 44}(2),
  533--546 (2017).

\bibitem{Moeskops2016}
Moeskops, P., Wolterink, J.~M., Van~der Velden, B. H.~M., Gilhuijs, K. G.~A.,
  Leiner, T., Viergever, M.~A., and I{\v{s}}gum, I., ``{Deep Learning for
  Multi-task Medical Image Segmentation in Multiple Modalities},'' in [{\em
  Lect. Notes Comput. Sci. (including Subser. Lect. Notes Artif. Intell. Lect.
  Notes Bioinformatics)}{\nolinebreak\hspace{0.1em}]},   {\bf 9901 LNCS},
  478--486, Springer, Cham (2016).

\bibitem{Amit2017}
Amit, G., Ben-Ari, R., Hadad, O., Monovich, E., Granot, N., and Hashoul, S.,
  ``{Classification of breast MRI lesions using small-size training sets:
  comparison of deep learning approaches},''  {\bf 10134},  101341H,
  International Society for Optics and Photonics (2017).

\bibitem{Elshof2010}
Elshof, L.~E., Rutgers, E. J.~T., Deurloo, E.~E., Loo, C.~E., Wesseling, J.,
  Pengel, K.~E., and Gilhuijs, K. G.~A., ``{A practical approach to manage
  additional lesions at preoperative breast MRI in patients eligible for breast
  conserving therapy: Results},'' {\em Breast Cancer Res. Treat.}~{\bf 124}(3),
   707--715 (2010).

\bibitem{Loo2011}
Loo, C.~E., Straver, M.~E., Rodenhuis, S., Muller, S.~H., Wesseling, J.,
  {Vrancken Peeters}, M.-J.~T., and Gilhuijs, K.~G., ``{Magnetic Resonance
  Imaging Response Monitoring of Breast Cancer During Neoadjuvant Chemotherapy:
  Relevance of Breast Cancer Subtype},'' {\em J. Clin. Oncol.}~{\bf 29}(6),
  660--666 (2011).

\bibitem{Dmitriev2013}
Dmitriev, I.~D., Loo, C.~E., Vogel, W.~V., Pengel, K.~E., and Gilhuijs, K.
  G.~A., ``{Fully automated deformable registration of breast DCE-MRI and
  PET/CT.},'' {\em Phys. Med. Biol.}~{\bf 58}(4),  1221--33 (2013).

\bibitem{VanderVelden2015}
Van~der Velden, B. H.~M., Dmitriev, I., Loo, C.~E., Pijnappel, R.~M., and
  Gilhuijs, K. G.~A., ``{Association between Parenchymal Enhancement of the
  Contralateral Breast in Dynamic Contrast-enhanced MR Imaging and Outcome of
  Patients with Unilateral Invasive Breast Cancer},'' {\em Radiology}~{\bf
  276}(3),  675--85 (2015).

\bibitem{Tustison2010}
Tustison, N.~J., Avants, B.~B., Cook, P.~A., Zheng, Y., Egan, A., Yushkevich,
  P.~A., and Gee, J.~C., ``{N4ITK: Improved N3 bias correction},'' {\em IEEE
  Trans. Med. Imaging}~{\bf 29}(6),  1310--1320 (2010).

\bibitem{Gilhuijs1998}
Gilhuijs, K.~G., Giger, M.~L., and Bick, U., ``{Computerized analysis of breast
  lesions in three dimensions using dynamic magnetic-resonance imaging.},''
  {\em Med. Phys.}~{\bf 25}(9),  1647--54 (1998).

\bibitem{Morris2013}
Morris, E., Comstock, C., and Lee, C.,  [{\em {ACR BI-RADS Magnetic Resonance
  Imaging}}{\nolinebreak\hspace{0.1em}]}, American College of Radiology,
  Reston, VA (2013).

\bibitem{Yu2016}
Yu, F. and Koltun, V., ``{Multi-Scale Context Aggregation by Dilated
  Convolutions},'' in [{\em ICLR}{\nolinebreak\hspace{0.1em}]},  (2016).

\bibitem{Wolterink2017}
Wolterink, J.~M., Leiner, T., Viergever, M.~A., and I{\v{s}}gum, I., ``{Dilated
  convolutional neural networks for cardiovascular MR segmentation in
  congenital heart disease},'' in [{\em Lect. Notes Comput. Sci. (including
  Subser. Lect. Notes Artif. Intell. Lect. Notes
  Bioinformatics)}{\nolinebreak\hspace{0.1em}]},   {\bf 10129 LNCS},  95--102,
  Springer, Cham (2017).

\bibitem{Gubern-Merida2015}
Gubern-M{\'{e}}rida, A., Mart{\'{i}}, R., Melendez, J., Hauth, J.~L., Mann,
  R.~M., Karssemeijer, N., and Platel, B., ``{Automated localization of breast
  cancer in DCE-MRI},'' {\em Med. Image Anal.}~{\bf 20}(1),  265--274 (2015).

\bibitem{Vignati2014}
Vignati, A., Giannini, V., Carbonaro, L.~A., Bertotto, I., Martincich, L.,
  Sardanelli, F., and Regge, D., ``{A new algorithm for automatic vascular
  mapping of DCE-MRI of the breast: Clinical application of a potential new
  biomarker.},'' {\em Comput. Methods Programs Biomed.}~{\bf 117}(3),  482--8
  (2014).

\bibitem{VanDerVelden2016}
Van~der Velden, B. H.~M., Schmitz, A. M.~T., Loo, C.~E., and Gilhuijs, K.
  G.~A., ``{Abstract P4-02-07: Association between computer-derived features of
  the ipsilateral breast on DCE-MRI and the 70-gene signature in patients with
  invasive breast cancer},'' {\em Cancer Res.}~{\bf 76}(4) (2016).

\end{thebibliography}
\bibliographystyle{spiebib} % makes bibtex use spiebib.bst

\end{document}